\documentclass[11pt,a4paper]{article}

\usepackage{times}
\usepackage{epsfig}
\usepackage{graphicx}
\usepackage{amsmath}
\usepackage{amssymb}
\usepackage{authblk}
\usepackage[margin=0.5in]{geometry}
\DeclareMathOperator*{\argmin1}{argmin}

\usepackage[pagebackref=true,breaklinks=true,letterpaper=true,colorlinks,bookmarks=false]{hyperref}

\begin{document}

\title{Local Geometry Inclusive Global Shape Representation}

\author[1]{Somenath Das\thanks{somenath@uga.edu}}
\author[1]{Suchendra M. Bhandarkar\thanks{suchi@cs.uga.edu}}
\affil[1]{Department of Computer Science, The University of Georgia}
\renewcommand\Authands{ and }
\date{June 17, 2017}
\maketitle


\begin{abstract}
Knowledge of shape geometry plays a pivotal role in many shape analysis applications. In this paper we introduce a local geometry-inclusive global representation of 3D shapes based on computation of the shortest quasi-geodesic paths between all possible pairs of points on the 3D shape manifold. In the proposed representation, the normal curvature along the quasi-geodesic paths between any two points on the shape surface is preserved. We employ the eigenspectrum of the proposed global representation to address the problems of determination of region-based correspondence between isometric shapes and characterization of self-symmetry in the absence of prior knowledge in the form of user-defined correspondence maps. We further utilize the commutative property of the resulting shape descriptor to extract stable regions between isometric shapes that differ from one another by a high degree of isometry transformation.  We also propose various shape characterization metrics in terms of the eigenvector decomposition of the shape descriptor spectrum to quantify the correspondence and self-symmetry of 3D shapes. The performance of the proposed 3D shape descriptor is experimentally compared with the performance of other relevant state-of-the-art 3D shape descriptors.
\end{abstract}


\section{Introduction}\label{sec:intro}

In the field of shape analysis, the computation of an optimal global description of a 3D shape is critically dependent upon the underlying application. Local shape geometry is important for applications where it is essential to establish point-to-point correspondence between candidate shapes. On the other hand, applications based on shape similarity computation rely on a suitably formulated global metric to characterize shape similarity. Based on the objective(s) of the application and nature or modality of the underlying shape data/information (i.e., geometric, topological, etc.), 3D shape analysis applications can be broadly categorized as purely geometric, semantic or knowledge-driven~\cite{van2011survey}. However, a large number of 3D shape analysis applications that belong to these categories or lie within their intersections are based upon a fundamental problem, i.e., that of determining the correspondence between the 3D shapes under consideration. Typical examples of these applications include rigid and non-rigid shape registration ~\cite{gelfand2005robust,chang2008automatic}, shape morphing ~\cite{kraevoy2004cross}, self-symmetry detection ~\cite{gal2006salient}, shape deformation transfer ~\cite{sumner2004deformation}, 3D surface reconstruction ~\cite{pekelny2008articulated}, shape-based object recognition and retrieval~\cite{jain2007spectral}, to name a few. In each of the aforementioned applications, shape descriptors play a critical role in determining the necessary 3D shape correspondence. Depending on the nature of the application, 3D shape descriptors could be purely geometric and used to capture the local 3D geometry of the shapes whereas others may incorporate prior knowledge about the global 3D shape. Ideally, a 3D shape descriptor should demonstrate robustness to topological noise while simultaneously capturing the underlying invariant shape features that are useful in computing the correspondence between 3D shapes.

In this paper, we address an important problem, i.e., that of determining correspondence between isometric 3D shapes (i.e., 3D shapes that have undergone isometry deformation or transformation with respect to each other) \textit{without} using any prior knowledge about the underlying shapes. To this end, we propose a 3D shape descriptor based on estimation of the approximate geodesic distance between all point pairs on the 3D shape manifold. The proposed representation is used to address the computation of 3D self-symmetry, determination of correspondence between isometric 3D shapes and detection of the most stable parts of the 3D shape under varying degrees of isometry (i.e, non-rigid pose) transformation between shapes. Since the geodesics over a 3D shape manifold are defined as surface curves of constant normal curvature, they naturally encode the local surface geometry along the curve. On a discrete triangulated 3D surface mesh, the discrete approximation to a geodesic is characterized by an optimal balance in terms of the angular distribution on either side of the discrete geodesic computed over the local neighborhood of each mesh point on the geodesic (Figure~\ref{fig:discrete-geodesics}). This balance of local angular distribution is observed to encode the local geometry of the triangulated mesh along the discrete geodesic. The aforementioned approximation to a geodesic computed over a discrete 3D triangulated mesh is referred to as a \textit{quasi-geodesic}~\cite{martinez2005computing}. The proposed global shape descriptor represents the 3D shape by computing the quasi-geodesic paths between all point pairs on the discrete 3D triangulated surface mesh. 

\begin{figure*}
\begin{center}
   \includegraphics[width=0.95\linewidth , keepaspectratio]{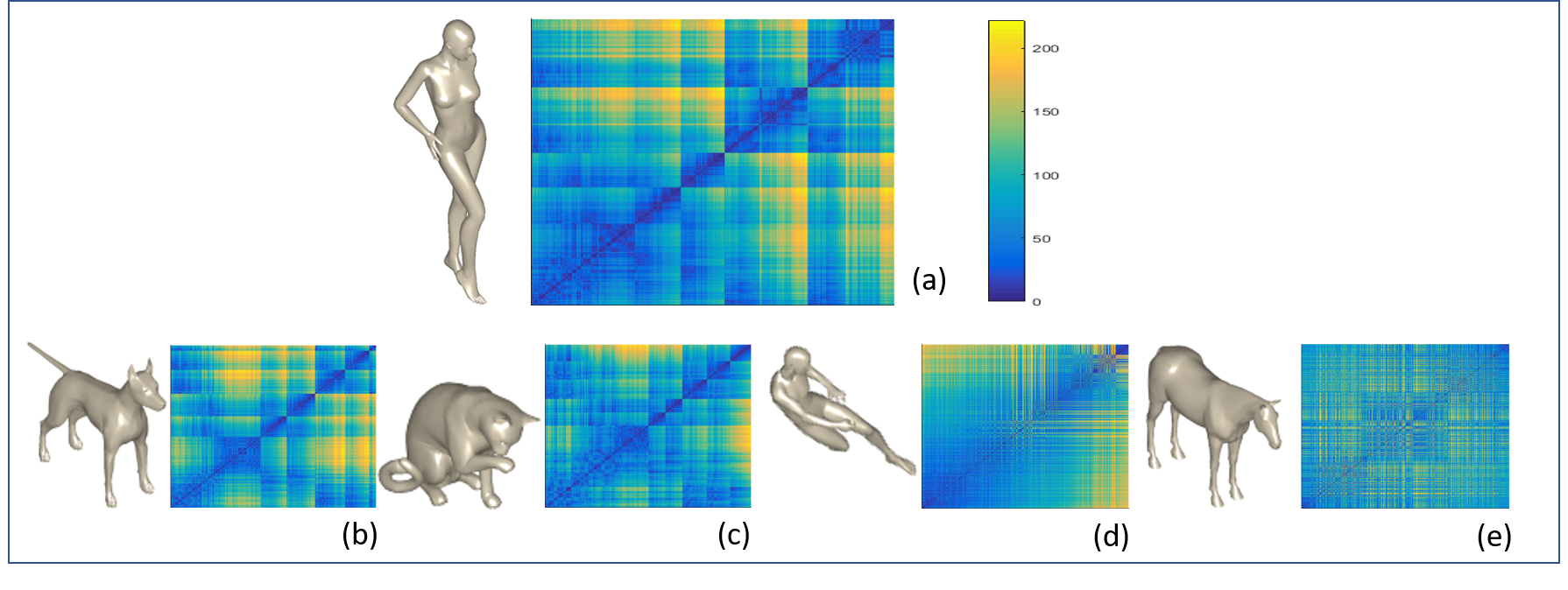}
\end{center}
 \caption{Global representation of 3D shapes using quasi-geodesics computed over a discrete triangulated 3D surface mesh. The 3D shape models shown are (a) {\it Victoria} (b) {\it Dog} (c) {\it Cat} (d) {\it Michael} and (e) {\it Horse}. The all-point-pairs quasi-geodesic matrix representation of the 3D shapes is observed to be approximately symmetric and the resulting eigenspectrum is observed to preserve self-symmetry over the discrete triangulated 3D mesh-based representation of the 3D shapes.}
\label{fig:2D-symm-maps}
\end{figure*}

The all-point-pairs geodesic matrix representation of 3D shapes presents a symmetrical pattern as shown in Figure~\ref{fig:2D-symm-maps}. We employ the eigenspectrum of this representation to detect self-symmetry within a shape. Furthermore, we investigate the commutative property of the eigenvectors of the shape descriptor spectrum, which are shown to be approximately orthogonal to each other for discrete setting such as triangulated mesh representation of shapes. Approximate orthogonality refers to the fact that two distinct eigen vector $\phi_i$ and $\phi_j$ of the spectrum would produce $<\phi_i . \phi_j> \approx 0$ when operated by a dot product operator $<.>$.
The proposed eigenspectral representation, however is distinct from the well known Laplace-Beltrami eigenspectrum that has been used extensively in several 3D shape analysis and 3D shape synthesis applications. In our case, we exploit the spectrum obtained by commuting the shape descriptor eigenspectrum to establish correspondence between isometric 3D shapes. It needs to be emphasized here that, unlike many related approaches~\cite{kovnatsky2013coupled,ovsjanikov2012functional}, the proposed optimization criterion used to establish the correspondence between isometric 3D shapes does not exploit nor does it require prior user-specified maps between the 3D shapes.

We use the proposed correspondence establishment scheme to test the hypothesis that the presence of implicit isometry between 3D shapes can be characterized using a global quasi-geodesic-based shape representation that encodes local shape geometry. Furthermore, we also contend that the proposed representation can be exploited to address problems such as self-symmetry detection and characterization, correspondence determination and stable part detection under isometry deformation without resorting to prior knowledge from external sources. To the best of our knowledge, the problem of correspondence determination in the absence of prior knowledge had not been addressed in the research literature. In some of our experiments, due to the high variability in the isometry transformations, we obtain poor results for correspondence determination as a result of not requiring any prior knowledge. However, our experiments show that the proposed correspondence determination technique is able to detect stable corresponding parts between shapes, i.e.,  corresponding parts that have undergone the least degree of isometry deformation (Section~\ref{sec:results}).

The remainder of the paper is organized as follows. In Section~\ref{sec:related-work}, we present a brief survey of the most relevant works on 3D shape description, 3D symmetry detection and characterization and 3D correspondence determination with an added emphasis on functional maps between isometric shapes~\cite{ovsjanikov2012functional}. Section~\ref{sec:contribution} describes the specific contributions of our work. The mathematical model on which the proposed technique is based is detailed in Section~\ref{sec:shape-operator}. In Section~\ref{sec:results}, we present the experimental results for 3D self-symmetry detection, 3D correspondence determination between isometric shapes, and stable 3D part detection. We conclude the paper in Section~\ref{sec:conclusion} with an outline of directions for future work.

\section{Related Work and Background}\label{sec:related-work}

The proposed global shape descriptor is based on the computation of all possible quasi-geodesics between all pairs of points over the discrete triangulated 3D surface mesh. The proposed descriptor is also capable of encoding the local geometry at the discrete points over the shape. As mentioned previously, the optimization criterion for establishing correspondence using the proposed descriptor does not require user specified prior knowledge of corresponding points on the candidate shapes. In this section, we first present a brief survey of relevant local shape representation schemes and spectrum-dependent shape correspondence models~\cite{heider2011localsurvey,van2011survey}. We also discuss relevant work on quasi-coupled harmonic bases~\cite{kovnatsky2013coupled}, which exploits the commutativity of the isometric shape spectrum to establish correspondence between approximately isometric shapes.

\subsection{Local shape descriptors}\label{sec-geom-desc}

The different classes of local shape descriptors can be categorized based on their approach towards sampling of the underlying local surface geometry. \textbf{Ring-based descriptors} are typically based on local sampling of a predefined metric over the discrete 3D surface mesh, however, they differ in their strategies for evaluation of the metric. Some of the prominent descriptors belonging to this class employ \textit{blowing bubbles}~\cite{shp-des-blowbubble-Mortara-ESV-Algca-2004,shp-des-blowbubbleExtension-Pottman-ESV-2009} centered around a sample surface point, whereas others use the geodesic diameter to sample the surface metric in a local neighborhood~\cite{pottmann2009integral}. These descriptors explicitly control the radius parameter of the bubbles or discs which in turn determines the size of the sample surface region. Some ring-based descriptors~\cite{gatzke2005curvature} use the local surface normal vectors computed at discrete points on the surface mesh to capture the local surface features. \textit{Geodesic fan descriptors}~\cite{shp-des-geofans-Gatzke-SMA-2005,shp-des-geonorm-Ong-2010} sample a local surface metric based on values of the local surface mesh curvature or the outward surface normal vector within regions of varying radii defined over the 3D surface mesh. \textit{Splash descriptors} employ the values of surface normal vector as the primary metric for local surface characterization~\cite{stein1992structural} whereas \textit{point descriptors}~\cite{shp-des-point-Chua-IJCV-1997,shp-des-pnt-Yamany-ICCV-1999,shp-des-pnt-Yamany-PAMI-2002} encode the local geometric features on the surface mesh defined by the relative local surface normal at a sample point with respect to a superimposed plane or line segment at the sample points. One of the more prominent examples from this category of shape descriptors is the point descriptor proposed by Kokkinos et al.~\cite{kokkinos2012intrinsic} where feature points are represented by local geometric and photometric fields.  \textbf{Expanding descriptors} fit a hypothesis-based model to a surface region in order to characterize it. Salient shape descriptors from this category typically employ a parametric model involving features such as geodesic distance~\cite{shp-des-blowbubble-Mortara-ESV-Algca-2004,shp-des-fitpoly-Cipriano-VCG-2009}, volume and/or surface area~\cite{shp-des-art-Connoly-JMG-2005,shp-des-blowbubbleExtension-Pottman-ESV-2009}. Some variants of this descriptor use a mesh smoothing~\cite{li2005multiscale} or mesh saliency computation~\cite{shp-des-art-Lee-TOG-2005} procedure that is employed over a specific region on the 3D surface mesh. \textbf{Iterative operator-based descriptors} capture the geometric changes within a shape by manipulating the entire mesh surface. As a manipulation strategy they employ techniques such as smoothing~\cite{shp-des-itrSmoothing-Li-Guskov-SOGP-2005} or estimation of local diffusion geometry~\cite{bronstein2010gromov} over the mesh surface. The well known Laplace-Beltrami operator~\cite{rustamov2007laplace} is typically employed to compute the diffusion-based shape descriptors.

\subsection{Global shape representation}\label{sec-globl-shp-rprsnt}

In most situations, knowledge of local surface geometry alone is insufficient to characterize the entire shape.  Consequently, a global shape representation based upon local surface features is necessary to effectively address the correspondence problem, which is fundamental to many applications in computer vision and computer graphics. In recent times, surface descriptors based on the eigenspectrum of the Laplace-Beltrami operator have gained popularity in the context of the correspondence problem. Some important examples of surface descriptors from this class are based on the formulation of a diffusion process. The diffusion process is guided by the Laplace-Beltrami operator~\cite{rustamov2007laplace} that samples a surface metric, such as the mesh connectivity, along the geodesic curves on the 3D surface mesh. In related work, Bronstein et al.~\cite{bronstein2010gromov} use diffusion geometry to measure the point-to-point length along a specific path on the surface mesh using a random walk model. Surface descriptors based on the heat kernel signature (HKS)~\cite{sun2009concise,bronstein2010scale} employ the heat diffusion model in conjunction with the eigenspectrum of the Laplace-Beltrami operator to characterize global shape. In an anisotropic variation, Boscaini et al.~\cite{boscaini2016anisotropic} use the eigenspectrum of a directional version of the Laplace-Beltrami operator for shape representation. The wave kernel signature (WKS)~\cite{aubry2011wave} is another popular category of shape descriptors based on the Laplace-Beltrami eigenspectrum, that employs the principles of quantum mechanics instead of heat diffusion to characterize the shape. Smeets et al.~\cite{smeets2012isometric} address the global representation of shape by computing the geodesic distances between sample points on the 3D surface mesh resulting in a shape representation scheme that is shown to achieve robustness against nearly isometric deformations.

\subsection{Joint diagonalization of the commutative eigenspectrum}\label{sec-joint-diag} 

Point or region specific correspondence between isometric shapes can be addressed by exploiting the commutative property of shape spectrum representation. In this section we briefly mention coupled quasi-harmonic bases ~\cite{kovnatsky2013coupled} that employs this commutative property of isometric (or near isometric) shape spectrum to address correspondence between isometric shapes.

\noindent
\textbf{Commutative Eigenspectrum}. Formally commutative property implies that, given two Unitary (Hermitian or orthogonal) operators $\Phi_X$ and $\Phi_Y$ defined over isometric shape pair $X$ and $Y$, there will be a joint diagonalizer $\Psi$ that would diagonalize both $\Psi^{T}\Phi_X\Psi$ and $\Psi^{T}\Phi_Y\Psi$ ~\cite{coisperturbation}. The joint diagonalizer $Psi$ would represent the common eigen bases between isometric shape spectrum $\Phi_X$ and $\Phi_Y$. Shapes represented as discrete triangulated mesh need not to be exactly isometric to each other due to existing discretization error. Therefore, for discrete case, the respective shape spectrums would be approximately commutative.  The term "approximately commutative" has been used in the following sense. Spectrums $\Phi_X$ and $\Phi_Y$ of triangulated shapes $X$ and $Y$ are approximately commutative if $||\Phi_X\Phi_Y - \Phi_Y\Phi_X||_F \approx 0$ where $||.||_F$ represents the Frobenius norm of matrix. A detailed treatment of common bases for approximately commutative spectral operators can be found in ~\cite{coisperturbation,yeredor2002non}. Some recent importatnt works ~\cite{kovnatsky2015functional,kovnatsky2013coupled} employ this principle to minimize an optimization criteria in least square sense to extract common spectrum bases that is used to address correspondence between isometric shapes. Specifically, we mention Coupled quasi-harmonic bases by Kovnatsky et. al. ~\cite{kovnatsky2013coupled} as follows.

\noindent
\textbf{Coupled Quasi-harmonic Bases} address correspondence between two approximately isometric shapes $X$ and $Y$ by finding common bases existing within the spectrum of two isometric or approximately isometric shapes. The proposed optmization criteria presented in the paper finds bases $\widehat{\Phi}_X$ and $\widehat{\Phi}_Y$ that jointly diagonalize the Laplacians $\Delta_X$ and $\Delta_Y$ defined over near isometric shapes $X$ and $Y$. The proposed optimization criteria extracts the common eigen bases $\widehat{\Phi}_X$ and $\widehat{\Phi}_Y$ by minimizing the optimization criteria in eqn. ~\ref{eq-quasi-opt}.

\begin{align}
\begin{split}
\argmin1_{\widehat{\Phi}_X,\widehat{\Phi}_X} \text{off}(\widehat{\Phi}_X^T W_X \widehat{\Phi}_X) + \text{off}(\widehat{\Phi}_Y^T W_Y \widehat{\Phi}_Y) + \\ ||F^T \widehat{\Phi}_X - G^T \widehat{\Phi}_Y||_F^2 \\ \text{such that }  \widehat{\Phi}_X^T D_X \widehat{\Phi}_X  = I \text{ and } \widehat{\Phi}_Y^T D_Y \widehat{\Phi}_Y = I\label{eq-quasi-opt}
\end{split}
\end{align}

Here $\text{off}(A) = \sum_{1\leq i \neq j \leq n} |a_{ij}^2|$ for an $n \times n$ matrix $A$ with elements $a_{ij}$. Matrices $W$ and $D$ are components of discrete cotangent laplacians $\Delta_X$ and $\Delta_Y$ for discrete meshes $X$ and $Y$ such that $\Delta_X = W_X^{-1} D_X$ and $\Delta_Y = W_Y^{-1} D_Y$ following the cotangent discretization of mesh Laplacian by Meyer et. al. ~\cite{meyer2003discrete}. The third component of optimization ~\ref{eq-quasi-opt} corresponds to coarse correspondence between $X$ and $Y$ provided a priori i a format of point-wise mapping stored within matrices $F$ and $G$ respectively.

In the present work we employ the common eigen bases between isometric shape spectrum to establish correspondence between them as well. However, in contrast with ~\cite{kovnatsky2013coupled} our optimization criteria does not exploit any prior correspondence information.

\section{Contributions of the Paper}\label{sec:contribution}

In this paper, we propose a global shape representation $D_g(X)$ for a 3D manifold $X$ that incorporates local surface geometry. The proposed representation is based on the computation of the shortest quasi-geodesic distances between all point pairs on the shape manifold. The proposed shape representation is shown to preserve the local surface geometry at each point on the 3D surface mesh. Furthermore, we effectively exploit the eigenspectrum of the proposed shape representation in the following applications: 


\noindent 
(1) {\it Self-symmetry characterization}: We address the problem of self-symmetry characterization by exploiting the eigenspectrum of the proposed global shape descriptor $D_g(X)$.

\noindent 
(2) {\it Correspondence determination}: We determine the region-wise correspondence between isometric shapes without requiring the user to determine and specify {\it a priori} the point-wise mapping between the two shapes.

\noindent 
(3) {\it Isometry deformation characterization}: We exploit the results of the region-wise correspondence to characterize and quantify the extent of isometry deformation between the shapes.

\noindent 
(4) {\it Stable part detection:} We exploit the commutative property of the eigenfunctions of $D_g(X)$ to extract pose-invariant stable parts within non-rigid shapes. 

\section{Local Geometry Inclusive Shape Operator}\label{sec:shape-operator}

In the proposed scheme a discrete 3D shape manifold $X$ is characterized by an operator $D_g(X)$, that is computed by determining the quasi-geodesics over the discrete manifold $X$. It is known that along a geodesic over a continuous manifold, only the normal component of the local curvature is dominant compared to the tangential component. A discrete 3D shape manifold $X$, in the form of a triangulated 3D surface mesh, can be represented by a $C^2$ differentiable function $f:\mathbb{R}^3\rightarrow\mathbb{R}$ as $X = \{f(x_{1}),f(x_{2}),...,f(x_{n})\}$ where $n$ denotes the number of vertices $x_i, 1 \leq i \leq n$ of $X$~\cite{azencot2014functional,martinez2014smoothed}. The quasi-geodesic computed for a discrete path $x_i\rightsquigarrow x_j$ minimizes the distance measure $d(f(x_i),f(x_j))$ between the vertices $x_i$ and $x_j$ of $X$. The proposed shape representation $D_g(X)$ records all such quasi-geodesics, computed between all vertex-pairs or point-pairs over the surface mesh $X$. Furthermore, the matrix representation of $D_g(X)$ reveals an implicit symmetrical form, as is evident for the example 3D shapes as shown in Figure~\ref{fig:2D-symm-maps}.

\begin{figure}[h]
\begin{center}
   \includegraphics[width=0.45\linewidth]{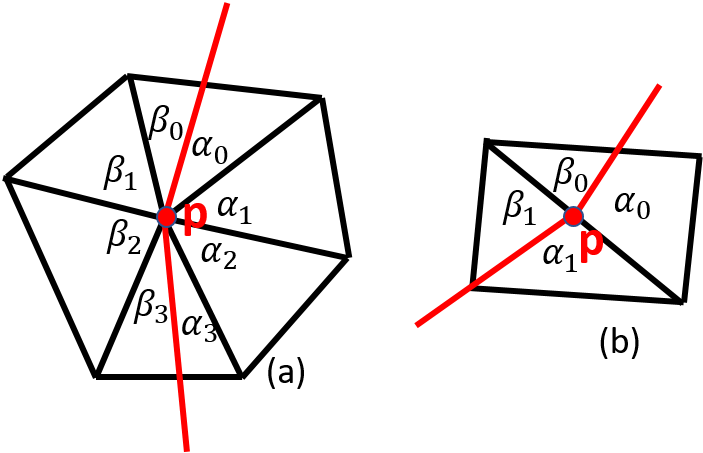}
\end{center}
   \caption{The right and left angular distributions $\theta_l$ and $\theta_r$ generated by a geodesic at point $P$ on the surface mesh. The angular measures $\theta_l$ and $\theta_r$ encode the local geometry on a discrete surface mesh.}
\label{fig:discrete-geodesics}

\end{figure}

For discrete meshes, the computation of geodesics is enabled by stable schemes such as ones described in~\cite{martinez2005computing}. The local geometry along a quasi-geodesic over a discrete mesh is preserved as follows:  Figure~\ref{fig:discrete-geodesics} depicts two scenarios where a probable quasi-geodesic (marked in red) crosses a neighborhood of triangular meshes. In either case, one can measure the discrete geodesic curvature at a point $P$ as follows: 
\begin{align}
\kappa_{g}(P) = \frac{2\pi}{\theta}(\frac{\theta}{2} - \theta_r) \label{eq-kappa-g}
\end{align}

\noindent
In eqn.~(\ref{eq-kappa-g}),  $\theta$ denotes the sum of all angles incident at point $P$ where the geodesic crosses the surface mesh. 
In both cases, depicted in Figure~\ref{fig:discrete-geodesics}(a) and (b), the quasi-geodesics create angular distributions $\theta_{l}$ and $\theta_{r}$ such that $\theta_l = \sum_i \beta_{i}$ and $\theta_r = \sum_i \alpha_{i}$. Since  the normal curvature is dominant along the quasi-geodesics, we can compute an optimum balance between $\theta_{l}$ and $\theta_{r}$ that minimizes the discrete geodesic curvature $\kappa_{g}$, which is the tangential curvature component along the quasi-geodesic. This optimal balance between angular distribution along the quasi-geodesic approximately encodes the local angular distribution as depicted in Figure~\ref{fig:discrete-geodesics}(a) and (b). 

The spectral decomposition of the symmetric shape operator $D_g(X)$ results in the eigenspectrum $\Phi_X$ for shape $X$ as follows: 
\begin{align}
D_g(X)\Phi_X = \Delta_X\Phi_X \label{eq-eigen}
\end{align} 
\noindent
where $\Delta_X = {\rm diag}(\gamma_1,\gamma_2,...,\gamma_n)$ denotes the diagonal matrix of eigenvalues $\gamma_i, 1 \leq i \leq n$ and $\Phi_X = \{\Phi_X^1, \Phi_X^2, ...,\Phi_X^n\}$ denotes the eigenvectors $\Phi_X^i, 1 \leq i \leq n$ of $X$ with $n$ surface vertices.

\subsection{Self-symmetry characterization}\label{sec:self-symm}

We characterize self-symmetric regions over shape $X$ as follows. Two regions $X_1, X_2 \subset X$ are possible candidates for being symmetric regions if for some upper bound $\varepsilon$: 
\begin{align}
\left |\sum_{k=1}^{k_0}\Phi_X^k(p)-\sum_{k=1}^{k_0}\Phi_X^k(q) \right |_2 \leq \varepsilon \quad \forall p\in X_1, \; \forall q\in X_2
\label{eq-symm}
\end{align}
\noindent
where $| \cdot |_2$ denotes the $\mathcal{L}^2$ norm. Using spectral analysis one can find a tight bound on $\varepsilon$ such that $\varepsilon \leq \sum_{p,q\in X_1, \; r,s \in X_2}\left |d(p,q)-d(r,s) \right |_2$ for a $C^2$ distance metric $d$ ~\cite{dunford1963linear}. Parameter $\varepsilon$  depends upon the variance of geodesic error computed over entire shape manifold $X$. Therefore, $\varepsilon$ for shape manifold  $X$ is a measure of the degree of isometry deformation of $X$ vis-a-vis the baseline shape. We report the bounds computed for different meshes in the Experimental Results section (Section~\ref{sec:results}). For characterizing self-symmetry we restrict ourselves to the lower-order eigenvectors denoted by $k_0 \leq 20$. Furthermore, the above characterization can be used to jointly analyze the correspondence between two candidate isometric shapes $X$ and $Y$ (Section~\ref{sec:correspondence}).

\subsection{Correspondence determination between isometric shapes}\label{sec:correspondence}

Determining the compatibility between the eigenbases of various shapes plays a critical role in applications dealing with analysis of multiple 3D shapes; in particular, determining the correspondence between 3D shapes. In related work, Ovsjanikov et al.~\cite{ovsjanikov2012functional} represent the correspondence between two shapes by a parametric map between their functional spaces. However, functional map-based methods typically rely on user-specified prior knowledge of the mapping between the shapes for optimization of the correspondence criterion~\cite{nguyen2011optimization,ovsjanikov2012functional}. In contrast, the proposed approach does not assume any user-specified prior mapping between the shapes under consideration. 

For correspondence determination between two isometric shapes $X$ and $Y$ we exploit the fact that the eigendecomposition of symmetric shape operators $D_g(X)$ and $D_g(Y)$ leads to approximately commutative eigenspectra $\Phi_X$ and $\Phi_Y$. The characterization "approximately commutative" is due to the discrete triangulation of meshes and follows the formal definition given in section ~\ref{sec-joint-diag}.  
 We couple $\Phi_X$ and $\Phi_Y$  by the commutative terms $\Phi_X^T \Delta_Y \Phi_Y$ and $\Phi_Y^T \Delta_X \Phi_X$ to solve the following optimization problem: 
\begin{align}
\bar{\Phi}_X , \bar{\Phi}_Y = \argmin1_{\phi_x, \phi_y} \quad |\phi_x^T \Delta_Y \phi_y|_F + |\phi_y^T \Delta_X \phi_x|_F
\label{eq-corr}
\end{align}
\noindent
where $\phi_x \subset \Phi_X$, $\phi_y \subset \Phi_Y$ and $| \cdot |_F$ denotes the Frobenius norm. It should be emphasized that eqn.~(\ref{eq-corr}) does not require that {\it a priori} correspondence maps be provided. The optimized maps $\bar{\Phi}_X$ and $\bar{\Phi}_Y$ over shapes $X$ and $Y$ encode the correspondence between them. From the optimized maps $\bar{\Phi}_X$ and  $\bar{\Phi}_Y$, the relative correspondence error between shapes $X$ and $Y$ is given by $C_{X,Y} = \sum_{k=1}^{k_0} |\bar{\Phi}_X^k - \bar{\Phi}_Y^k|_2$. To compute $C_{X,Y}$ we consider the lower-order eigenvectors by setting $k_0 \leq 20$. 

\subsection{Stable 3D region or part detection}\label{sec:stable-part}

Relaxing the criterion for correspondence determination by not requiring a user-specified prior mapping between the shapes could result in poor  correspondence between shapes that differ significantly from each other via isometry transformation. However, the optimization criterion for correspondence determination can be also used to identify common stable regions or parts within the shapes. The stable regions or parts are deemed to be ones that have undergone the least amount of isometry deformation as a result of pose variation. We present the following criterion to identify the stable regions $S_{X,Y}$ between shapes $X$ and $Y$ as follows.
\begin{align}
S_{X,Y} = \bigcup_p \; |\bar{\Phi}_X(p) - \bar{\Phi}_Y(p)|_2 \leq \varepsilon
\label{eq-stable}
\end{align}
\noindent
where region $p$ represents a corresponding region in both shapes $X$ and $Y$ as identified by the correspondence optimization in eqn.~(\ref{eq-corr}). The parameter $\varepsilon$ is computed as mentioned in Section~\ref{sec:self-symm}. The stable part detection is quantified using the following criterion: $\bar{S}_{X,Y} = \sum_{p \in S_{X,Y}} \quad |\bar{\Phi}_X(p) - \bar{\Phi}_Y(p)|_2$.

\section{Experimental Results} \label{sec:results}

\begin{figure}[h]
\begin{center}
   \includegraphics[width=0.45\linewidth , keepaspectratio]{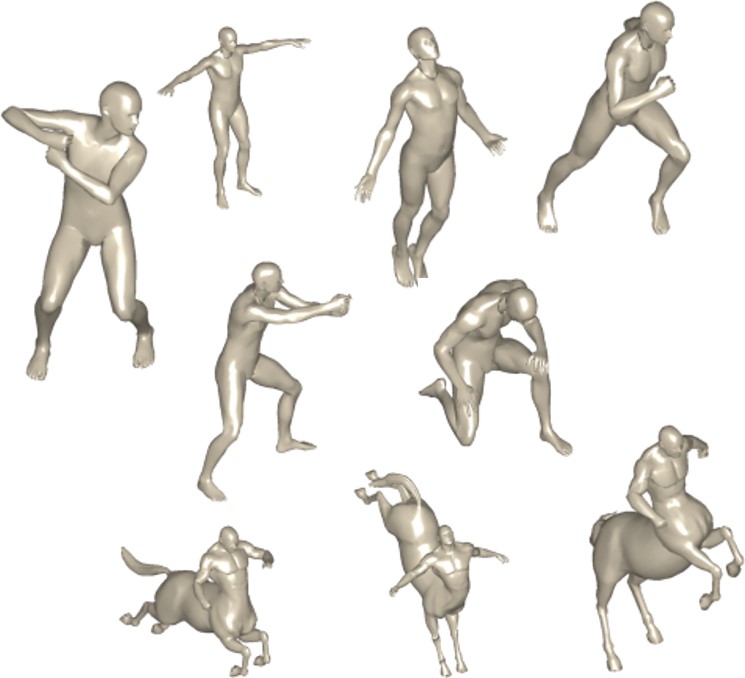}
\end{center}
   \caption{Examples of isometry transformation for the shape categories {\it human} and {\it centaur} in the TOSCA dataset.}
\label{fig:example-tosca}
\end{figure}

For our experiments we have chosen the TOSCA dataset consisting of ten non-rigid shape categories, i.e., {\it Cat}, {\it Dog}, {\it Wolf}, two {\it Human Males}, {\it Victoria}, {\it Gorilla}, {\it Horse}, {\it Centaur} and {\it Seahorse}~\cite{ovsjanikov2009shape}. Within each shape category, the individual shapes represent different transformations such as isometry, isometry coupled with topology change, different mesh triangulations of the same shape etc. In this work, we consider shapes that are isometric to one another, i.e., shapes that differ via an isometry transformation. Examples of some shapes that differ from one another via isometry transformations are shown in Figure~\ref{fig:example-tosca}. Experimental results are presented for six different shape categories for each of the applications formally described in Sections~\ref{sec:self-symm},~\ref{sec:correspondence} and~\ref{sec:stable-part} using visual representations of the results followed by the corresponding numerical evaluations. We have experimented with coarse meshes that are reduced by more than 87\% of their original size or resolution. The results of the proposed shape representation are compared with those from relevant state-of-the-art shape representation schemes. The comparable performance achieved by the proposed local geometry-inclusive global shape representation scheme without requiring any prior knowledge of point-to-point or region-wise correspondence validates the central hypothesis behind the proposed scheme that the implicit isometry existing within candidate shapes can be employed to establish correspondence without the knowledge of coarse correspondence provided a priori.

\subsection{Results of 3D self symmetry detection}\label{sec-res-sym}

\begin{figure*}
\begin{center}
   \includegraphics[width=0.75\linewidth]{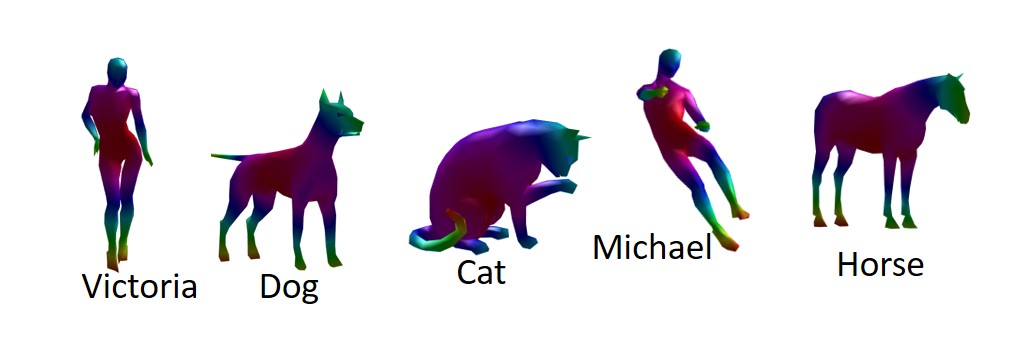}
\end{center}
   \caption{Self-symmetry detection for five different TOSCA shape categories using the spectrum of the global representation $D_g(X)$ for the shape $X$. Each map corresponds to the second eigenvector $\Phi_X^2$ of the shape operator spectrum.}
\label{fig:self-symm}
\end{figure*}

Figure~\ref{fig:self-symm} depicts the self-symmetry maps obtained for the various shapes using eqn.~(\ref{eq-symm}). The maps in 
Figure~\ref{fig:self-symm} correspond to the second eigenvector $\Phi_X^2$ obtained from the spectral decomposition of the global operator $D_g(X)$ for each shape using eqn.~(\ref{eq-eigen}). Table~\ref{tab-epsilon} presents the self-symmetry characterization measure, denoted by the upper bound $\varepsilon$ in eqn.~(\ref{eq-symm}), for each shape category. This characterization measure represents the average degree of isometry transformation within a shape category vis-a-vis the baseline shape. Note that the shape category {\it Michael} represents one of the two {\it Human Male} shape categories in the TOSCA dataset.

\begin{table}[h]
\begin{center}
\begin{tabular}{|c|c|c|c|}
\hline
Category & $\varepsilon$ & Category & $\varepsilon$\\
\hline
{\it Victoria} & 0.528 & {\it Dog} & 0.462 \\
\hline
{\it Cat} & 0.282 & {\it Michael} & 0.566 \\
\hline
{\it Horse} & 0.815 & {\it Centaur} & 0.203\\
\hline
\end{tabular}
\end{center}
\caption{Self-symmetry characterization measure for different shape categories in the TOSCA dataset. The average degree of isometry transformation within the category {\it Horse} is observed to be at least 30\% higher than the others.}
\label{tab-epsilon}
\end{table}

\subsection{Results of 3D correspondence between isometric shapes}\label{sec-res-corr}

\begin{figure*}
\begin{center}
   \includegraphics[width=0.65\linewidth , keepaspectratio]{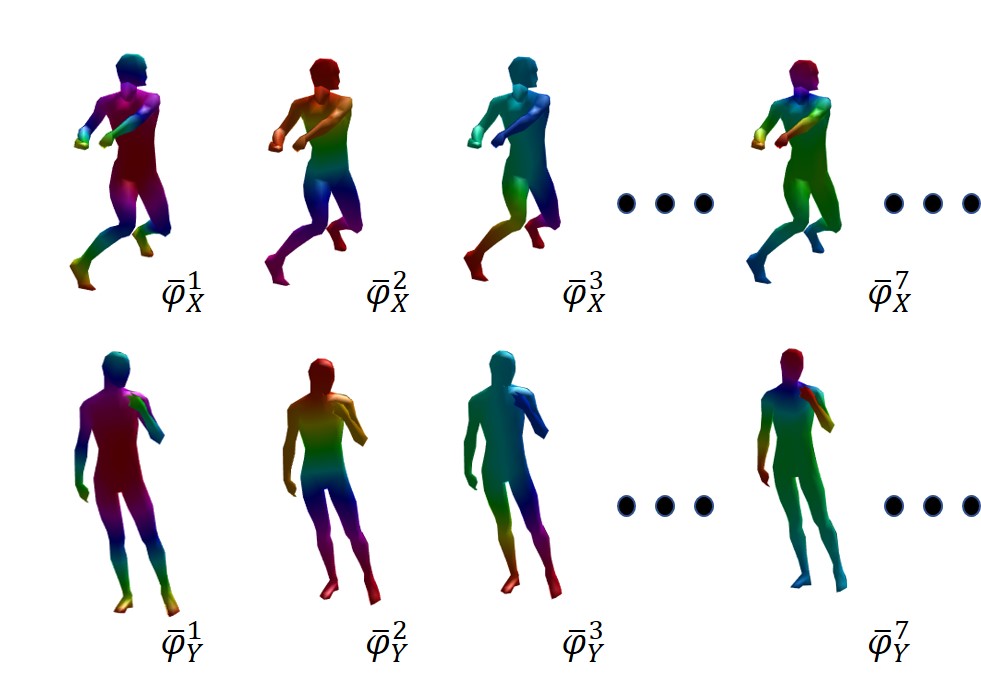}
\end{center}
   \caption{Pairwise consistency between corresponding eigenmaps of the {\it Human Male} shapes. For correspondence estimation, the optimization procedure described in eqn.~(\ref{eq-corr}) is used. Lower-order eigenvectors are considered for correspondence estimation since they effectively capture the global shape geometry.}
\label{fig:corr-consistency}
\end{figure*}

Since the lower-order eigenvectors represent global shape geometry more accurately, we consider the first 20 eigenvectors to compute the global region-based correspondence between the isometric shapes. Figure~\ref{fig:corr-consistency} shows the results of correspondence determination between the isometric {\it Human Male} shapes obtained via the optimization procedure described in eqn.~(\ref{eq-corr}). The correspondence maps between the shapes are shown to be consistent across the different order eigenvectors.

\begin{table}[h]
\begin{center}
\begin{tabular}{|c|c|c|c|}
\hline
Category & Average $C_{X,Y}$ & Category & Average $C_{X,Y}$\\
\hline
{\it Victoria} & 0.069 & {\it Dog} & 0.0624 \\
\hline
{\it Cat} & 0.06 & {\it Michael} & 0.057 \\
\hline
{\it Horse} & 0.0559 & {\it Centaur} & 0.052\\
\hline
\end{tabular}
\end{center}
\caption{Average relative error $C_{X,Y}$ in 3D correspondence determination between isometric shapes.}
\label{tab-cxy-corr}
\end{table}

The relative correspondence error for these maps can be characterized by the measure $C_{X,Y}$ defined in Section~\ref{sec:correspondence}. Table~\ref{tab-cxy-corr} lists this measure for isometric shapes from different TOSCA shape categories. Lower $C_{X,Y}$ values denote a higher degree of correspondence accuracy achieved via the optimization procedure described in eqn.(~\ref{eq-corr}). Once again, we emphasize that the correspondence accuracy is achieved without requiring any user-specified prior mapping between the shapes. 

\subsection{Results of 3D stable region or part detection}

\begin{figure}[h]
\begin{center}
   \includegraphics[width=.45\linewidth]{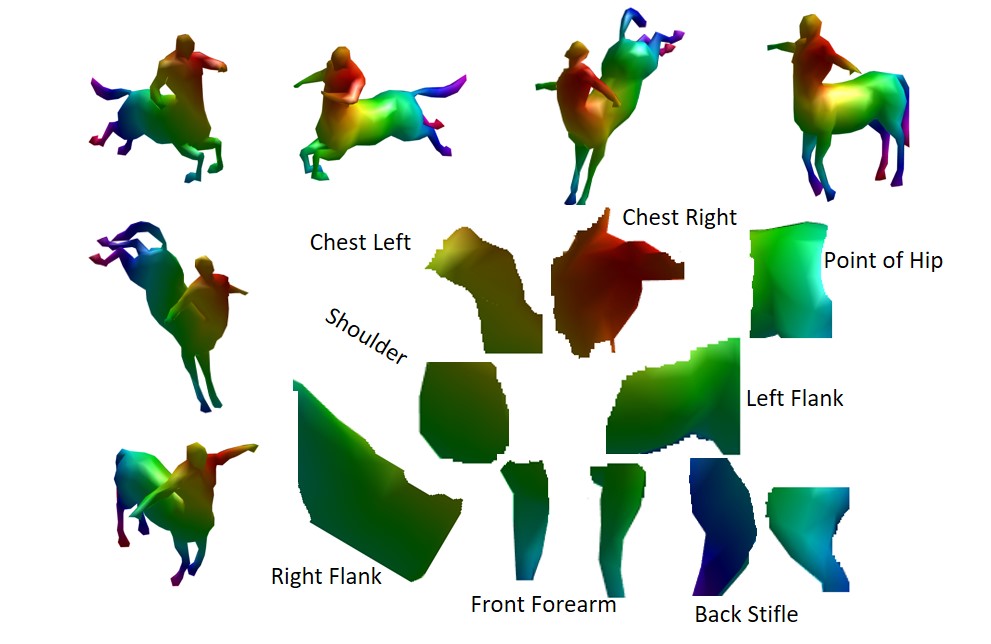}
\end{center}
   \caption{Stable region detection via the optimization procedure described in eqn.~(\ref{eq-stable}). The stable regions are detected between isometric shapes where the correspondence accuracy is observed to deteriorate due to a high degree of isometry transformation between the shapes. Unstable regions are ones that exhibit a higher degree of isometry transformation between them, for example, parts of the lower legs, the tail,  etc.}
\label{fig:stable-region-detection}
\end{figure}

Shapes from different categories display varying degrees of isometry transformations between them. As a result, the accuracy of global correspondence deteriorates for shapes that exhibit a very high degree of isometry deformation. This is expected since the proposed scheme does not assume any prior mapping information that could potentially improve the correspondence. However, using the optimization described in eqn.~(\ref{eq-stable}) we can identify the stable corresponding regions or parts within the shapes that are least transformed by isometry. The detected stable regions or parts for the {\it Centaur} shape category are depicted in Figure~\ref{fig:stable-region-detection}. For various poses of the {\it Centaur} shape model, the more dynamic regions such as the tail and the lower legs exhibit low correspondence accuracy and hence are rejected by the optimization criterion in eqn.~(\ref{eq-stable}). However, regions that are least affected by the isometry deformation are detected as stable regions. These stable regions exhibit high correspondence accuracy and are depicted in Figure~\ref{fig:stable-region-detection}. We quantify the correspondence accuracy for the detected stable regions using the measure $\bar{S}_{X,Y}$ described in Section~\ref{sec:stable-part}. However, in our experiments, we observed a high positive correlation between measures $C_{X,Y}$ and $\bar{S}_{X,Y}$. Hence we state that the results in Table~\ref{tab-cxy-corr} hold for measure $\bar{S}_{X,Y}$ as well.

Table~\ref{tab-comparison} compares the performance of the proposed representation scheme with the performance of other state-of-the-art representation schemes \cite{kim2011blended,sahillioglu2011coarse}. These methods were further combined with functional map ~\cite{ovsjanikov2012functional} to improve their  correspondence through functional map supported local refinement. Performance comparison of these combined approaches with the proposed representation is also presented in Table~\ref{tab-comparison}. The numerical values presented in Table~\ref{tab-comparison} represent the highest percentage correspondence accuracy achieved by the various representation schemes along with the corresponding average geodesic errors.  The performance of the proposed representation scheme compares very well with the performance of the other state-of-the-art representation schemes. We emphasize here on the merit of the cetral hypothesis behind the present shape representation that, the comparable performance in self-symmetry detection, and correspondance map between isometric shapes is achieved by the proposed representation scheme without using any prior correspondence mapping information between the shapes unlike the other state-of-the-art correspondence models \cite{kim2011blended,sahillioglu2011coarse}.

\begin{table}[h]
\begin{center}
\begin{tabular}{|c|c|c|}
\hline
Methods & Geodesic Error & \% Correspondence\\
\hline
\cite{kim2011blended} & 0.11 & $\sim 95$\\
\hline
~\cite{ovsjanikov2012functional} and \cite{kim2011blended} & 0.06 & $\sim 95$\\
\hline
\cite{sahillioglu2011coarse} & 0.25 & $\sim 90$\\
\hline
\cite{ovsjanikov2012functional} and~\cite{sahillioglu2011coarse} & 0.2 & $\sim 90$\\
\hline
{Proposed Scheme} & 0.15 & $\sim 94$\\
\hline
\end{tabular}
\end{center}
\caption{Comparison between the proposed scheme and those described in Kim et al.~\cite{kim2011blended}, Ovsjanikov et al.~\cite{ovsjanikov2012functional} and Sahillioglu and Yemez~\cite{sahillioglu2011coarse} and their combinations.}
\label{tab-comparison}
\end{table}

\section{Conclusions and Future Work}\label{sec:conclusion}

In this paper we proposed a global shape representation scheme using quasi-geodesics computed over the entire discrete shape manifold. The spectral decomposition of this representation is effectively used to identify self-symmetric regions on the discrete shape manifold. By exploiting the commutative property of the eigenbases of the proposed representation, we successfully demonstrated its use in correspondence determination between isometric shapes. We also proposed characterization metrics for self-symmetry identification and correspondence determination. Furthermore, stable regions within shapes were identified for shape pairs that differ from each other by a high degree of isometry deformation. The results of correspondence determination obtained via the proposed representation scheme were compared with those from relevant state-of-the-art representation schemes. 

The key contribution of this work is the fact that no prior knowledge, in the form of user-specified mappings, was used for correspondence determination or self-symmetry detection. As an extension of the present scheme, we intend to explore and combine functional maps~\cite{ovsjanikov2012functional} that may prove critical in exploring the group structure within isometric shapes. Furthermore, we intend to use this combined scheme to address correspondence determination between near-isometric shapes~\cite{kovnatsky2013coupled}.

{\small
\bibliographystyle{ieee}
\bibliography{./iccv_cefrl_17}
}

\end{document}